\documentclass[manuscript,screen]{acmart}
\usepackage{array}
\usepackage{multirow}

\newcolumntype{P}[1]{ >{\centering\arraybackslash} m{#1}}

\AtBeginDocument{%
  \providecommand\BibTeX{{%
    \normalfont B\kern-0.5em{\scshape i\kern-0.25em b}\kern-0.8em\TeX}}}

\copyrightyear{2022}
\acmYear{2022}
\setcopyright{acmcopyright}\acmConference[HT '22]{Proceedings of the 33rd ACM Conference on Hypertext and Social Media}{June 28-July 1, 2022}{Barcelona, Spain}
\acmBooktitle{Proceedings of the 33rd ACM Conference on Hypertext and Social Media (HT '22), June 28-July 1, 2022, Barcelona, Spain}
\acmPrice{15.00}
\acmDOI{10.1145/3511095.3531268}
\acmISBN{978-1-4503-9233-4/22/06}

\begin{document}

\title[Sentence Specific Popularity Forecasting]{Towards Proactively Forecasting Sentence-Specific Information Popularity within Online News Documents}


\author{Sayar Ghosh Roy}
\email{sayar.ghosh@research.iiit.ac.in}
\orcid{0000-0002-9885-8181}
\affiliation{
  \institution{IIIT Hyderabad}
  \streetaddress{Professor CR Rao Road, Gachibowli}
  \country{India}
  \postcode{500032}
}

\author{Anshul Padhi}
\email{anshul.padhi@research.iiit.ac.in}
\orcid{0000-0001-9620-9850}
\affiliation{
  \institution{IIIT Hyderabad}
  \streetaddress{Professor CR Rao Road, Gachibowli}
  \country{India}
  \postcode{500032}
}

\author{Risubh Jain}
\email{risubh.jain@students.iiit.ac.in}
\orcid{0000-0001-5216-9973}
\affiliation{
  \institution{IIIT Hyderabad}
  \streetaddress{Professor CR Rao Road, Gachibowli}
  \country{India}
  \postcode{500032}
}

\author{Manish Gupta}
\email{gmanish@microsoft.com}
\orcid{0000-0002-2843-3110}
\affiliation{
  \institution{IIIT Hyderabad}
  \streetaddress{Professor CR Rao Road, Gachibowli}
  \country{India}
  \postcode{500032}
}

\affiliation{  
  \institution{Microsoft}
  \streetaddress{Campus, Gachibowli}
  \country{India}
  \postcode{500032}
}

\author{Vasudeva Varma}
\email{vv@iiit.ac.in}
\orcid{0000-0003-1923-1725}
\affiliation{
  \institution{IIIT Hyderabad}
  \streetaddress{Professor CR Rao Road, Gachibowli}
  \country{India}
  \postcode{500032}
}

\renewcommand{\shortauthors}{Ghosh Roy et al., 2022}


\begin{abstract}
Multiple studies have focused on predicting the prospective popularity of an online document as a whole, without paying attention to the contributions of its individual parts. We introduce the task of proactively forecasting popularities of sentences within online news documents \textit{solely} utilizing their natural language content. We model sentence-specific popularity forecasting as a sequence regression task. For training our models, we curate \textsc{InfoPop}, the \textit{first} dataset containing popularity labels for over 1.7 million sentences from over 50,000 online news documents. To the best of our knowledge, this is the first dataset automatically created using streams of incoming search engine queries to generate sentence-level popularity annotations. We propose a novel transfer learning approach involving sentence salience prediction as an auxiliary task. Our proposed technique coupled with a BERT-based neural model exceeds nDCG values of 0.8 for proactive sentence-specific popularity forecasting. Notably, our study presents a non-trivial takeaway: though popularity and salience are different concepts, transfer learning from salience prediction enhances popularity forecasting. We release \textsc{InfoPop} and make our code publicly available\footnote{{\url{https://github.com/sayarghoshroy/InfoPopularity}}}.
\end{abstract}



\maketitle

\section{Introduction}
In a typical document popularity prediction task, the objective is to estimate the would-be popularity of some content, say, a news article that is published on an online platform. A popularity prediction system might evaluate the number of pageviews that a particular online document would receive, or a social-media popularity prediction model might estimate the number of signatures, comments, shares, or likes that a specific social media post would accrue over a certain period of time.

Though popularity prediction is a well-studied machine learning task, with approaches ranging from traditional feature-based methods to recent Deep Neural Network architectures, all of the existing works on popularity prediction have focused \textit{exclusively} on the document-level prediction of prospective popularity labels~\cite{predict_news_headline_pop,news_article_pop}. When a system predicts that a specific article will be popular over time, a natural follow-up is to ask why. Which information pieces in the document would receive the most notice? Could we identify the informative sentences that would dominate the popularity level of the article? Being able to generate such insights \textit{proactively} would have wide-ranging business applications in article promotion, popularity-guided summarization and pull quote selection, etc.

But \textit{sentence-specific} forecasting of popularity has not been attempted, not even in a simplified binary classification of sentences into popular or not setting. The difficulty in obtaining reliable sentence-level labeled data for information popularity has been a crucial driving factor for the same.

An online article's popularity can be defined in terms of its page views~\cite{pop_headline_gen,forecast_pop_news}, which is a by-product of the regular Internet browsing activities of the global population. Similarly, within a particular online document, information popularity of a unit sentence is proportional to the number of received requests actively seeking its contained information. To capture this, we leverage incoming search queries encountered by a popular search engine over a sufficient period to ascertain sentence-specific popularity labels based on how queried-after each sentence is within a document. We present \textsc{InfoPop}, a dataset of over 50,000 online news articles with over 1.7M sentences, mapped to popularity scores.

\begin{figure}[t]
    \centering    
    \includegraphics[width=0.65\textwidth]{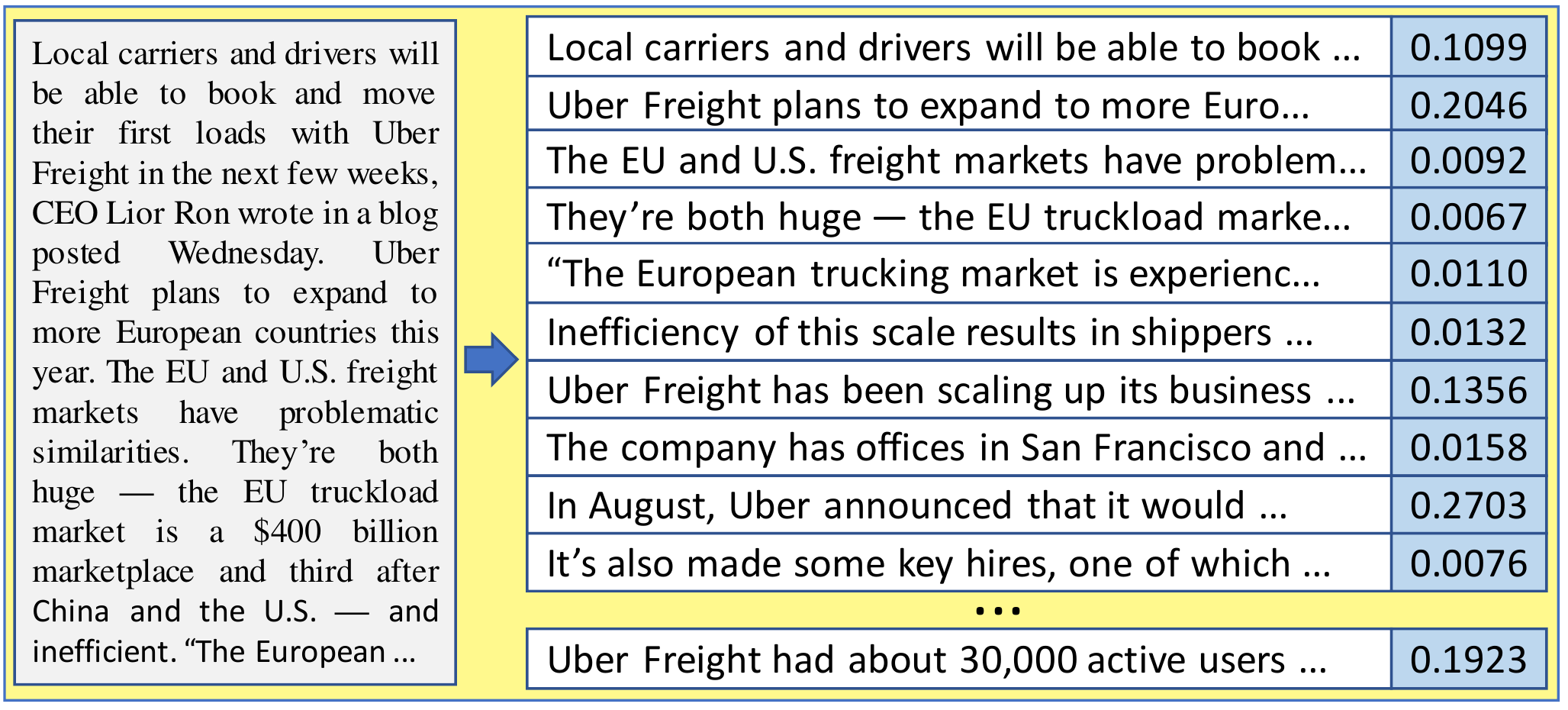}
    \caption{Task outline. Looking \textit{solely} at document text as input, forecast prospective sentence-specific popularity scores}
    \label{fig:task_outline}
\end{figure}

Taking inspiration from existing research on document-level popularity prediction, we frame the problem of sentence-specific popularity forecasting as a regression task similar to previous studies~\cite{predicting_pop,petition_pop} (who model document-level popularity) as opposed to a simple binary classification of a document's sentences into popular or not. 
Unlike some document-popularity forecasting approaches that rely on post-publication signals like pageview hits in the first half-hour after publication~\cite{predicting_pop,pop_time_ser}, we forecast sentence popularity scores \textit{proactively} (before a document's publication). Note that we utilize incoming search queries \textit{only} to define sentence-specific popularity and construct the \textsc{InfoPop} dataset. However, our problem formulation is one of \textit{query-insensitive} relative normalized scoring of sentences. Formally, given \textit{just} a document as input, our task (Fig.~\ref{fig:task_outline}) is of assigning normalized scores in the range $[0, 1]$, to every sentence indicating their intra-document relative information popularity, without utilizing any \textit{external} signals.

The key novelty of our approach lies in the design of auxiliary transfer learning subtasks adapting the STILTs (Supplementary Training on Intermediate Labeled-data Tasks)~\cite{stilts} technique. In brief, we create auxiliary subtasks of sentence \textit{salience prediction} in the domain of online news, framed under an identical problem formulation as that of our primary task of popularity forecasting (Sec.~\ref{sec:methods}). We experiment with a baseline neural model using RNNs (Recurrent Neural Networks) and a robust sequence regression architecture based on BERT (Bidirectional Encoder Representations from Transformers)~\cite{bert}, showcasing our proposed approach's efficacy. Also, we handle arbitrarily long documents with our neural models employing a sliding window mechanism over a document's sentences. In our analysis, we illustrate the differences in the concepts of popularity and salience, both qualitatively and quantitatively (Sec.~\ref{sec:results}). The empirically demonstrated efficacy of our approach brings out an interesting yet non-trivial takeaway -- though salience and popularity are varying concepts, transfer learning from salience prediction improves popularity forecasting.

Overall, in this paper, we make the following contributions.

\begin{itemize}
    \itemsep 0.8em
    \item We introduce the task of proactively forecasting relative information popularities of sentences within online news documents \textit{solely} based on their natural language content, without the use of any external features.
    \item We present \textsc{InfoPop}, the first labeled dataset containing over 50,000 online news documents from 26 news websites with over 1.7M sentences, each mapped to supervised popularity scores.
    \item Through a novel STILTs-based transfer learning approach, we build high-performance neural models reaching nDCG scores over 0.8 for sentence-specific popularity forecasting.
\end{itemize}

The remainder of the paper is organized as follows. In Section~\ref{sec:relatedWork}, we discuss related work on document-level popularity prediction, automatic text summarization and snippet generation. We formally present the concept of sentence-specific information popularity in Section~\ref{sec:sent_popl}. We describe details of curation of our dataset, \textsc{InfoPop} in Section~\ref{sec:infopop}. Next, we outline our proposed methods in Section~\ref{sec:methods}. Further, we discuss evaluation metrics in Section~\ref{sec:metrics}, experimental details in Section~\ref{sec:experiments} and present detailed results with analysis in Section~\ref{sec:results}. Finally, we conclude with a brief summary in Section~\ref{sec:conclusion}.

\section{Related Work}
\label{sec:relatedWork}
In this section, we discuss related work on document-level popularity prediction, automatic text summarization and snippet generation.

\subsection{Document-level Popularity Prediction}

Existing works in popularity prediction have considered a piece of content as an atomic unit whose prospective popularity is inferred. Based on the choice of popularity surrogate, we distinguish between two kinds of document popularity prediction problems. 

\begin{itemize}
    \itemsep1em
    
    \item \textbf{Popularity based on Internet browsing}: Studies on forecasting popularities of online news documents typically consider the number of received page load requests over sufficient time (accumulated pageviews or hit count) as a surrogate for article-level popularity~\cite{predict_news_headline_pop,forecast_pop_news,predicting_pop}.
    
    \item \textbf{Social Media Popularity}: These approaches aim to estimate the prospective engagement of a piece of content put up on a particular social media platform. Previous works have dealt with multi-modal posts~\cite{catboost}, images~\cite{temp_dyn_flickr}, movies~\cite{movie_pop}, petitions~\cite{petition_pop}, etc. and have utilized user-behavior based markers such as the number of comments~\cite{predict_mining,ranking_news_pop}, shares~\cite{pop_content_meta,multi_source_social}, etc. as surrogates of social-media popularity. There has also been limited work on time-aware~\cite{time-aware-movie} and time-series prediction of social-media popularity~\cite{time_pop,petition_time}. 

\end{itemize}

Unlike all previous approaches that determine popularity at the document-level, we focus on sentence-specific popularity forecasting (as shown in Fig.~\ref{fig:task_outline}) with our domain of information being online news, not social-media posts.

For \textit{informative} documents (including online news), a preferred surrogate of popularity has been the number of pageview hits~\cite{entity_pop,news_article_pop,pop_headline_gen}. Intuitively, pageviews captures the generic browsing trends of the population \textit{not limited} to social media actions. Our task, however, requires knowledge of more fine-grained Internet browsing activity for annotating specific sentences with popularity labels.

\subsection{Automatic Text Summarization}

Text summarization aims to recognize information that is central to the core idea of one or more document(s). Approaches are roughly divided into two categories, namely, abstractive: generate a concise, coherent, and cogent summary that captures the central idea expressed by the text piece~\cite{neural_abs_summ,lewis-etal-2020-bart}, and extractive: select and arrange salient plus diverse ranges of text to form a summary~\cite{hibert,text_match}. Extractive summarization is often framed as a sequence classification problem -- 
binary labels are assigned to sentences indicating their presence in the summary~\cite{cnn_dm_hermann,bertsumm}.
In contrast, our task necessitates sequential regression (numerical sentence-specific scores are forecasted). Also, text summarization has the additional objective of generating text \textit{without} repeated information. In that, an otherwise summary inclusion-worthy sentence may have a binary label of 0 if a similar sentence is already present in the oracle summary~\cite{distilsum} (in an extractive setting). However, we expect almost identical sentences in an article to have very similar information popularity values.

\subsection{Snippet Generation}

A page snippet can be a document excerpt that lets a user understand whether a document is pertinent to their query without accessing it in its entirety.
Snippet-generation~\cite{snippet_3} has a ranking-based problem formulation somewhat related to sentence popularity forecasting. However, our task is one of query-\textit{insensitive} scoring of sentences. In contrast, tasks such as snippet generation (and document retrieval~\cite{docret}) are query-\textit{sensitive}~\cite{snippet_q_2} -- typical methods require a particular query tailored to which an appropriate snippet is generated (except for na\"ive approaches that output initial tokens).

\section{Sentence-Specific Information Popularity}
\label{sec:sent_popl}
Document popularity magnitudes and popular-or-not labels based on pageviews capture the amount of notice that the document receives on the Internet over a period of time~\cite{forecast_pop_news,predict_news_headline_pop} and is a consequence of the everyday Internet browsing actions of the worldwide population.

\begin{table}[t]
\small
\centering
\caption{Selected sentences from a document in \textsc{InfoPop} with their true and forecasted popularities and predicted salience. Popularity forecasts are from our best performing model on nDCG (BERT$_{\scriptsize{\textnormal{Reg}}}$ with TL = S$_{\scriptsize{\textnormal{L}}}$). Salience predictions are based on BERT$_{\scriptsize{\textnormal{Reg}}}$ trained on S$_1$.\\[4pt]{[TPL: True Popularity Label, FPL: Forecasted Popularity Label, PSL: Predicted Salience Label, TPR: True Popularity Rank, FPR: Forecasted Popularity Rank, PSR: Predicted Salience Rank]}}
\label{tab:data_InfoPop}
\begin{tabular}{|p{9.2cm}|P{0.68cm}|P{0.68cm}|P{0.68cm}|P{0.38cm}|P{0.38cm}|P{0.38cm}|}
\hline
Sentence&TPL&FPL&PSL&TPR&FPR&PSR\\
\hline
\hline
Local carriers and drivers will be able to book and move their first loads with Uber Freight in the next few weeks, CEO Lior Ron wrote in a blog posted Wednesday. &0.1099&0.1522&0.1237&5&4&1\\
\hline
Uber Freight plans to expand to more European countries this year. &0.2046&0.1914&0.0924&2&2&4\\
\hline
The EU and U.S. freight markets have problematic similarities. &0.0092&0.0274&0.0736&10&6&7\\
\hline
They’re both huge -- the EU truckload market is a \$400 billion marketplace and third after China and the U.S. -- and inefficient.&0.0067&0.0048&0.0818&12&7&6\\
\hline
“The European trucking market is experiencing a severe shortage of drivers, and of the time drivers are on the road, 21 percent of total kilometers travelled are empty”, Ron wrote. &0.0110&0.0041&0.0892&9&8&5\\
\hline
Inefficiency of this scale results in shippers struggling to find available drivers to move their goods. &0.0132&0.0015&0.0530&7&12&10\\
\hline
Uber Freight has been scaling up its business since launching in May 2017, growing from limited regional operations in Texas to the rest of the continental U.S. &0.1356&0.1301&0.1018&4&5&2\\
\hline
The company has offices in San Francisco and Chicago. &0.0158&0.0024&0.0590&6&10&9\\
\hline
In August, Uber announced that it would make Uber Freight a separate unit and more than double its investment into the business.&0.2703&0.1968&0.0988&1&1&3\\
\hline
It’s also made some key hires, one of which intimated the company’s global ambitions.&0.0076&0.0015&0.0382&11&11&12\\
\hline
The company has made headway breaking into the U.S. market.&0.0116&0.0025&0.0675&8&9&8\\
\hline
Uber Freight had about 30,000 active users last quarter. &0.1923&0.1906&0.0427&3&3&11\\
\hline
\end{tabular}
\end{table}

With increased Internet penetration, the average number of queries encountered daily by commercial search engines has exceeded the billion mark\footnote{\url{https://searchengineland.com/google-now-handles-2-999-trillion-searches-per-year-250247}}. Google Trends\footnote{\url{https://trends.google.com}} showcases popular topics of interest segregated by region and timespan based on the pool of encountered search queries, and the collection of worldwide queries serves as a guide to mark information of universal interest. Correspondingly, within the local context of a single news document (say $D$), if an information piece, say $I_1$, is more queried-after by the global population than, say, piece $I_2$ in $D$, then: $popularity(I_1)$ $>$ $popularity(I_2)$. Popularity levels of sentences within $D$ can be derived based on this base principle, wherein the popularity of a specific sentence (say $s$) in $D$ is incremented based on the lexical similarity between $s$ and an encountered query.

For a news document $D$ with sentences [$s_1$, $s_2$, ..., $s_N$], comparing each sentence $\{s_i\}_{i=1}^N$ to \textit{every} search query encountered by a commercial search engine would be computationally infeasible. Moreover, only a negligibly small percentage of all encountered queries would positively contribute to sentences' popularities within $D$. Such queries would precisely be the ones for which $D$ could be deemed relevant. Thus, to derive scores of $s_i$'s within $D$, we filter incoming queries and consider the sublist, say $Q$ = [$q_1$, $q_2$, ..., $q_{|Q|}$], for which the document $D$ was \textit{significantly} relevant.

We regard $D$ as a \textit{significantly} relevant document for query $q$ if $D$ was shown within the top $10$ search results (which roughly translates to the first page of news results) when the search engine encountered $q$. We assign a base score of $\sum_{j=1}^{|Q|}similarity(q_j, s_i)$ to each sentence $\{s_i\}_{i=1}^N$. Sentence popularity scores are then normalized to a $[0, 1]$ range such that the popularity labels of sentences within $D$ sum up to $1$.

\section{\textsc{InfoPop} Dataset}
\label{sec:infopop}

In this section, we outline the creation of the \textsc{InfoPop} dataset and present its basic statistics.

\subsection{Data Acquisition}

\begin{figure}[t]
\includegraphics[width=0.85\columnwidth]{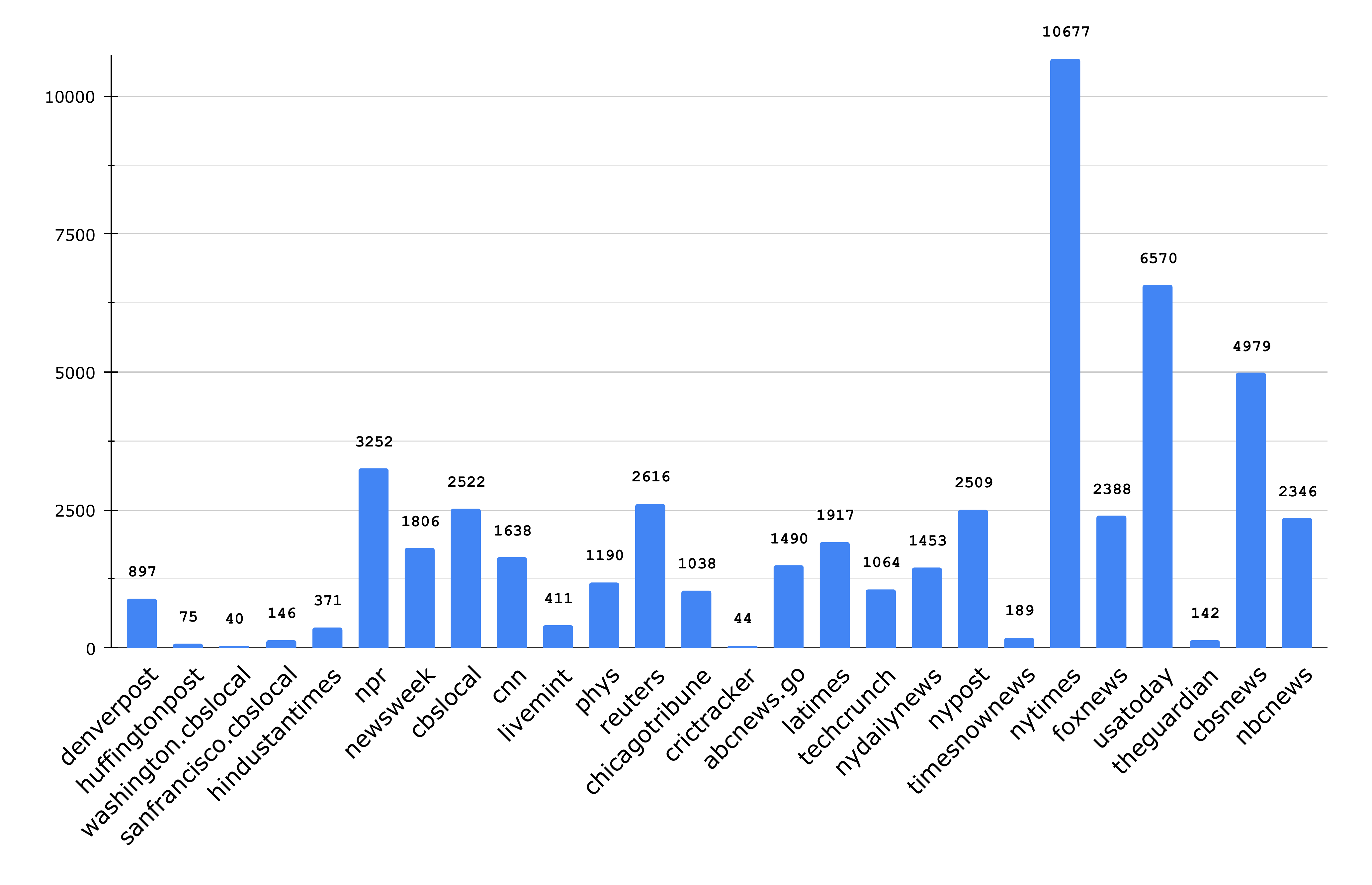}
\caption{\textsc{InfoPop}: news sources (on x-axis) with their corresponding \#documents (on y-axis)}
\label{fig:sources}
\end{figure}

We scraped 82,540 news documents from 26 reputed online news websites (websites listed in Fig.~\ref{fig:sources}). We accessed incoming queries of Microsoft Bing Search and mapped each document to the global assemblage of queries that deemed the document as \textit{significantly} relevant (Sec.~\ref{sec:sent_popl}) at the time when it was encountered. We extracted each article's text content and split that into an ordered list of sentences. With news articles as unit data points, individual sentences served as our atomic information pieces. We discounted articles with over 100 sentences as we found many of these to be long non-news documents.

\subsection{Preprocessing}
We encountered a considerable amount of noise in the extracted webpage texts. These primarily manifested as strings of seemingly random word tokens. We removed such heavily non-grammatical sentences using two dependency parsing-based heuristics. 
\begin{itemize}
    \itemsep0.5em
    \item Remove a sentence if its dependency graph is not a tree. The dependency graph of a grammatical sentence is supposed to be a tree and not a forest. We evaluated whether the produced dependency graph of a sentence was both connected and acyclic or not.
    \item Remove a sentence if its dependency parse contains some \textit{xcomp} branch leading to a single participle. The X complement or \textit{xcomp} is a dependency label used to mark dependent clauses that do not contain a subject. Consider a noisy unit such as `I would like to, I would like to, I would like to.' The parse tree of such a text string contains an \textit{xcomp} branch leading to a participle at the leaf node. We observed a positive correlation between such phenomena and text repetition.
\end{itemize}

Though the above cleaning heuristics are computationally expensive, we employ these owing to the one-time nature of the operation. After these cleanup steps, we further removed articles containing less than three grammatical sentences.

\subsection{Popularity Labels}
For a particular online document $D$ (say), we considered the collection of all queries, say $Q$ = [$q_1$, $q_2$, ..., $q_{|Q|}$], for which $D$ was \textit{significantly} relevant. We generated supervised popularity labels for every sentence in $D$ using the method outlined in Sec.~\ref{sec:sent_popl}. More formally, $P_i$ $\leftarrow$ $\frac{\sum_{j=1}^{|Q|}similarity(q_j, s_i)}{\sum_{i=1}^{N} \sum_{j=1}^{|Q|}similarity(q_j, s_i)}$, $\forall i \in \{1, 2, ..., N\}$ where $P_i$ denotes the normalized popularity label assigned to the $i^{th}$ sentence in $D$, assuming $D$ contains $N$ sentences: $s_1$, $s_2$, ..., $s_N$.

We used cosine-similarity between corresponding TF-IDF vectors as the measure of similarity. Though our similarity function might feel primitive, we uphold that utilizing lexical similarity is optimum in this case since search engine ranking algorithms still utilize word match as the basic principle and most relevant pages typically contain the query terms (that usually do not have well defined semantic embeddings) themselves. That said, we plan to explore semantic embedding based similarity as part of future work.

\begin{figure}[t]
\includegraphics[width=0.75\columnwidth]{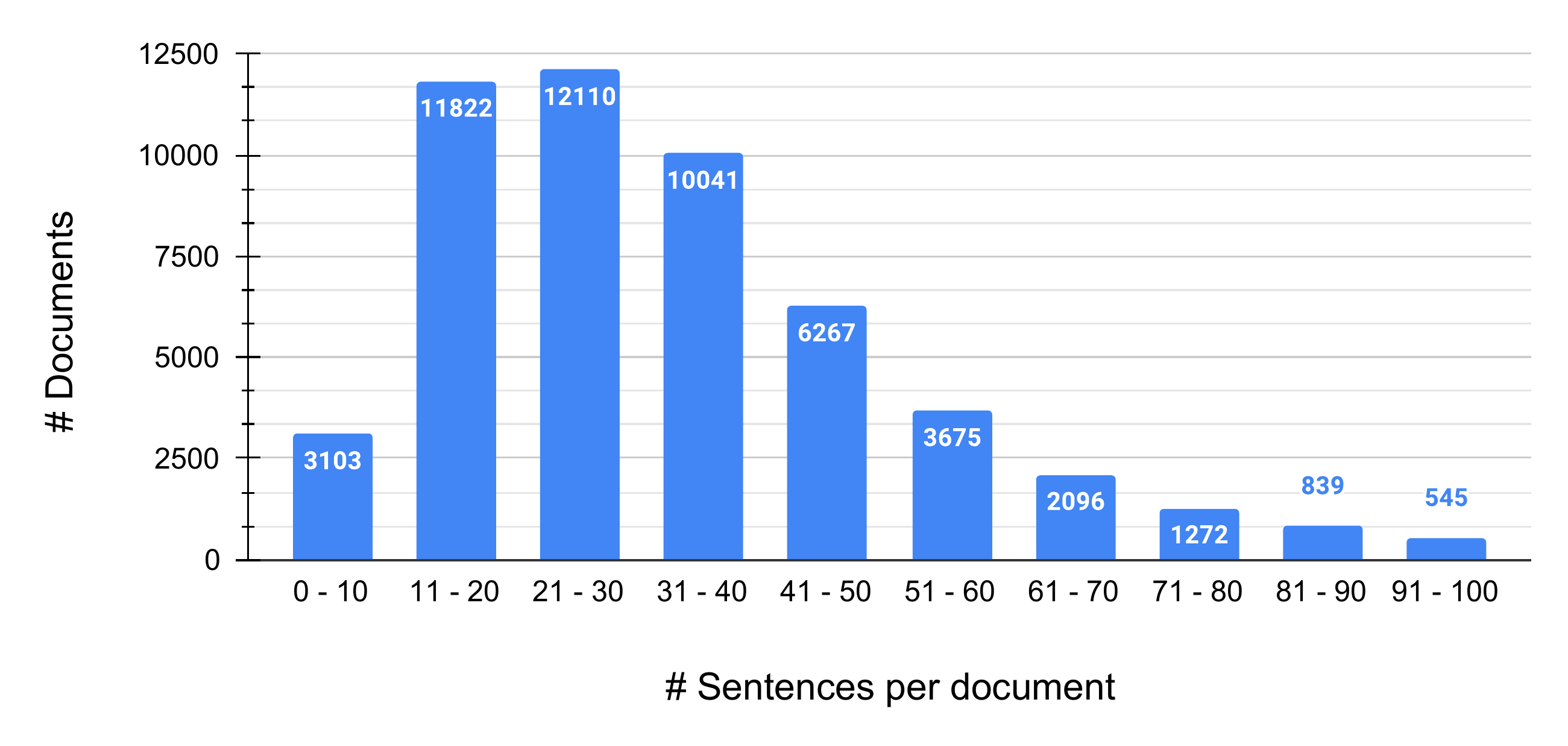}
\caption{\textsc{InfoPop}: Distribution of \#sentences per document
}
\label{fig:sent_distro}
\end{figure}

\subsection{\textsc{InfoPop} Dataset Statistics}

\textsc{InfoPop} contains 51,770 news documents containing 1,711,890 sentences annotated with normalized popularity scores. Each document contains a minimum of 3 and a maximum of 100 sentences, with the average number of sentences per article being 33.07. Fig.~\ref{fig:sent_distro} shows the distribution of the number of sentences per document.
On average, each sentence contains 18.23 word tokens. The minimum and maximum document lengths stand at 15 and 2,516 tokens, respectively, with 602.92 tokens being the average document length. The news articles are sourced from 26 reputed news websites, with the average number of articles per website being 1991.15. Fig.~\ref{fig:sources} illustrates the specific number of documents sourced from each news website.
We observe that popularity scores and sentence lengths have a weak positive correlation of 0.168. We divide our dataset into train, validation, and test splits in an 8:1:1 ratio.

\noindent To ensure reproducibility, we make \textsc{InfoPop} (and our defined train, validation, and test splits) publicly available here\footnote{\scriptsize{\url{https://github.com/sayarghoshroy/InfoPopularity}}}. Table~\ref{tab:data_InfoPop} shows an excerpt from a document within \textsc{InfoPop} with sentence-specific popularity labels (TPL).

\section{Methods}
\label{sec:methods}

We describe various methods for proactive sentence-specific popularity forecasting, including unsupervised sentence ranking baselines, neural sentence sequence regression architectures, and our proposed supervised Transfer Learning approach from an auxiliary task of salience prediction.

\subsection{Unsupervised sentence ranking baselines}

News articles typically utilize the pyramid structure of reporting with primary information contained within the initial sentences. We experiment with a position baseline where we score sentences in descending order of their position from the beginning of the article. Specifically, we assign the score of $1 - \frac{i}{n}$ to the $i^{th}$ sentence of an article with $n$ sentences. 
We also evaluate the performance of TextRank~\cite{textrank} and LexRank~\cite{LexRank} algorithms that exploit graph-based similarities between all sentence pairs through PageRank.

\subsection{Our Proposed Approach}
\label{sec:approach}

The primary novelty of our proposed approach lies in the design of a STILTs-based~\cite{stilts} Transfer Learning (TL) setup using a constructed task of sentence salience prediction. Recent advances in NLP have utilized transfer learning in two broad ways, (a) \textit{unsupervised} pre-training of text encoders using specific Language Modeling objectives~\cite{roberta,domain_adapt} and (b) \textit{supervised} pre-training of architectures on tasks with similar problem formulations~\cite{stilts,intermediate_TL}. In STILTs (Supplementary Training on Intermediate Labeled-data Tasks), a model pre-trained on unsupervised corpora (like BERT) is further pre-trained on an intermediate, supervised task for which ample labeled data is available before fine-tuning on its primary task.

We observe how sentence salience is well studied in context of text summarization, an area that enjoys availability of sizeable amounts of news domain data and hypothesize that STILTs-based transfer learning from an auxiliary task of salience prediction would boost our models' popularity forecasting ability.

\subsubsection{Auxiliary Transfer Learning (TL) Subtasks}

For document summarization, a sentence is considered salient if it contains information related to the primary semantics of the document and can be a worthy inclusion for the document's summary~\cite{summarunner}. Given an article and its gold standard summary, the salience of an individual sentence is computed as the ROUGE overlap between sentence and summary.
For oracle creation in a typical extractive summarization setting, binary summary-inclusion labels are computed for sentences \textit{greedily} to maximize the ROUGE overlap between the \textit{complete oracle} and the true summary~\cite{summarunner}. Such a greedy labeling scheme implicitly takes minimization of information redundancy into account. In that, a salient sentence might receive a summary-inclusion label of 0 if a lexically similar sentence was previously included in the oracle summary~\cite{distilsum}. However, we \textit{only} capture the salience of sentences and desire very similar sentences to have similar labels.
Similar to sentence-specific popularity forecasting, we package our auxiliary task as a sentence sequence regression problem. This lets us (a) train our architectures adapting the STILTs approach and (b) perform an empirical cross-task evaluation (Sec.~\ref{sec:cross_task}).

Since our domain of interest is online news, we utilize the publicly available CNN-DailyMail news summarization dataset (with the same splits as in \cite{cnn_dm_hermann}). We compute \textit{three} weakly supervised salience scores for each sentence based on its ROUGE 1, ROUGE 2, and ROUGE L overlap with the corresponding article's summary. Like \textsc{InfoPop} labels, we normalize salience labels across a document by dividing by the sum of individual sentence scores. Thus, we create three auxiliary subtasks due to the three labeling schemes, which we tag as S$_1$, S$_2$, and S$_{\scriptsize{\textnormal{L}}}$, respectively.

\begin{figure}[t]
\centering
\includegraphics[width=0.55\columnwidth]{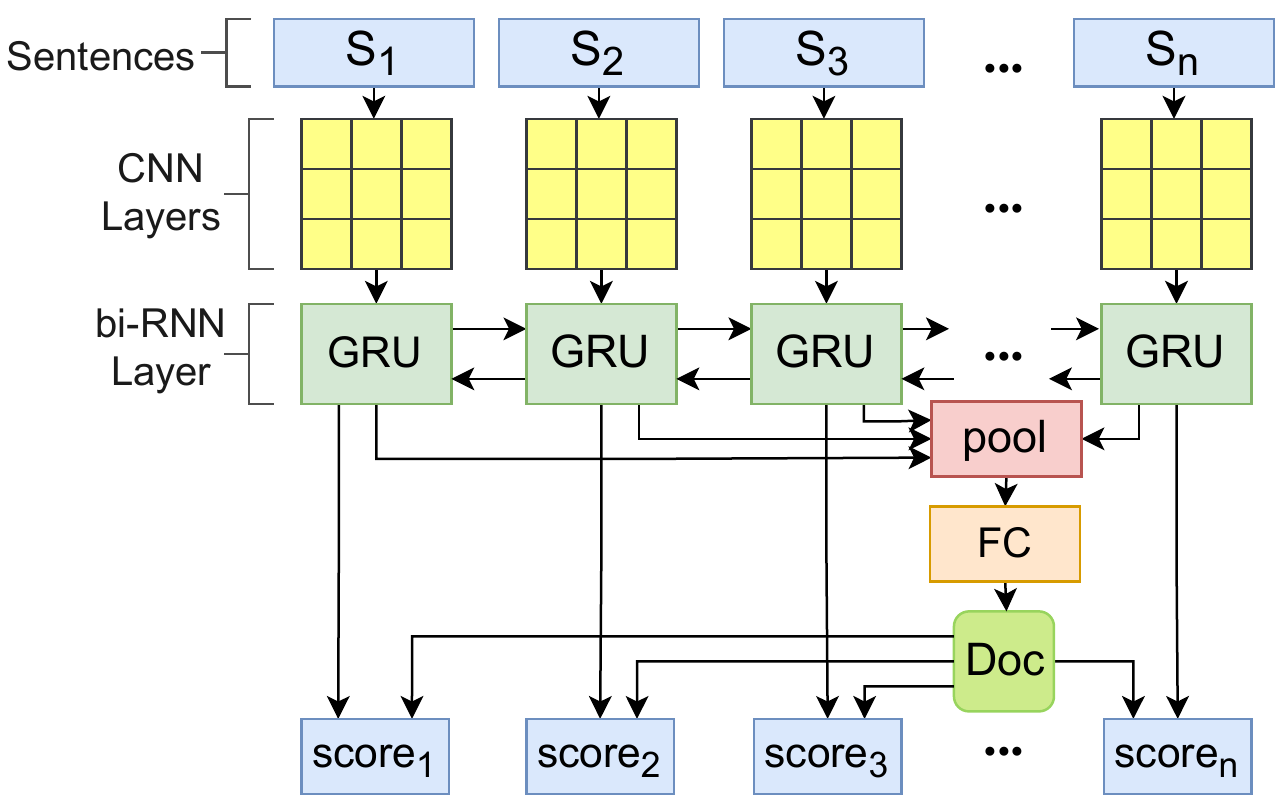}
\caption{Base$_{\scriptsize{\textnormal{Reg}}}$: RNN for Sentence Sequence Regression}
\label{fig:base_reg_condensed}
\end{figure}

\subsubsection{Supervised Neural Architectures}

\begin{figure}[t]
\includegraphics[width=0.58\columnwidth]{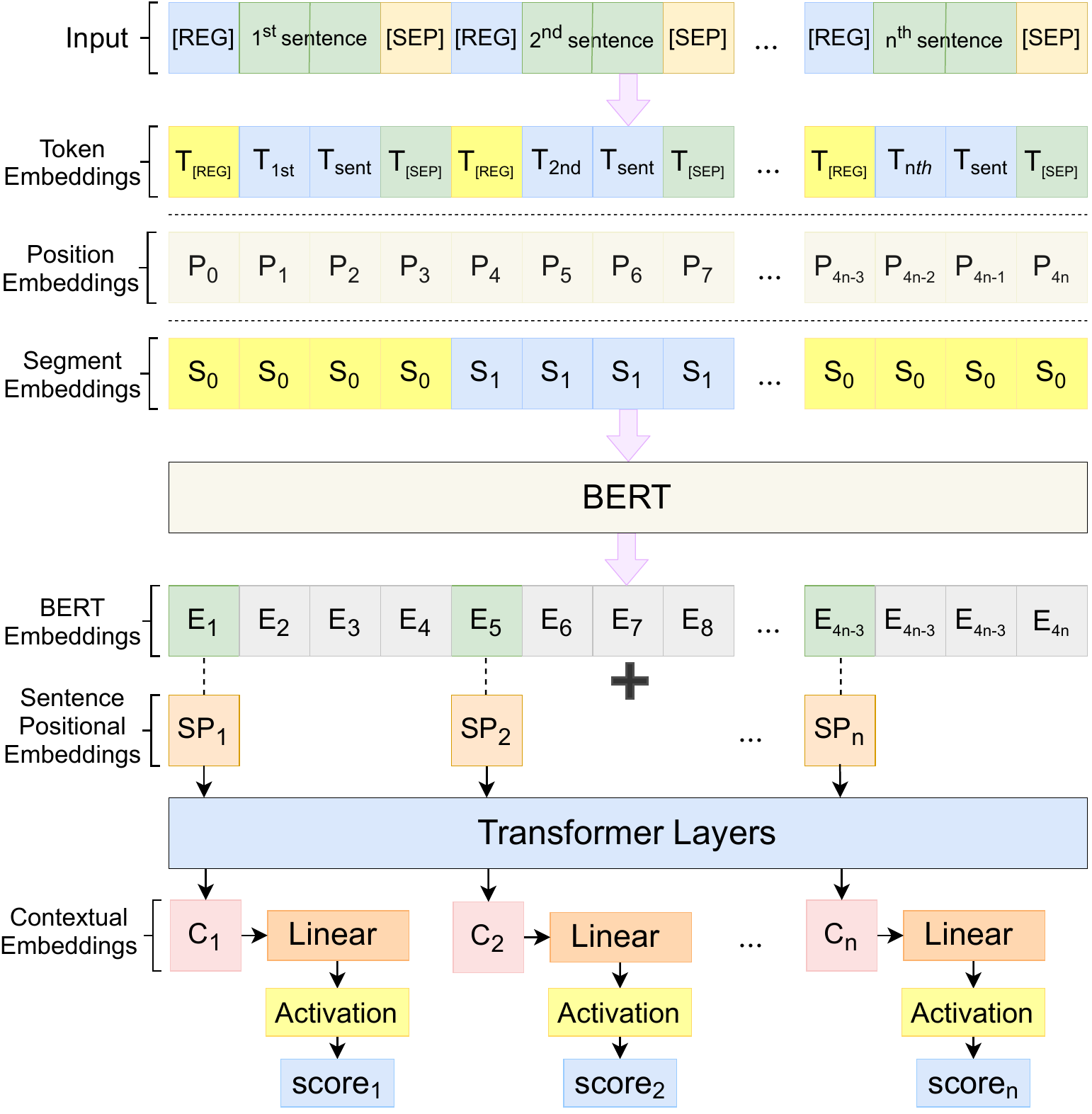}
\caption{BERT$_{\scriptsize{\textnormal{Reg}}}$: BERT for Sentence Sequence Regression}
\label{fig:bert_reg}
\end{figure}

We frame information popularity forecasting as a sentence sequence regression task where a sentence's score is relative to the entire article. Thus, effective neural models require the global context.
We pre-train a neural architecture on a supervised TL subtask (S$_1$, S$_2$, or S$_{\scriptsize{\textnormal{L}}}$), and then fine-tune the model for popularity forecasting. High-level overviews of Base$_{\scriptsize{\textnormal{Reg}}}$, our rudimentary neural baseline, and BERT$_{\scriptsize{\textnormal{Reg}}}$, our BERT-based sentence-sequence regression model are illustrated in Fig.~\ref{fig:base_reg_condensed} and Fig.~\ref{fig:bert_reg}, respectively. We use MSE (Mean Squared Error) loss between true and inferred sentence scores for training our neural models. Note that we handle arbitrarily large documents using both BERT$_{\scriptsize{\textnormal{Reg}}}$ and Base$_{\scriptsize{\textnormal{Reg}}}$ employing a sliding window mechanism over a document's sentences with a preset stride. \\



\noindent\textbf{Base$_{\scriptsize{\textnormal{Reg}}}$}: As a simple baseline neural model for sentence scoring, we build upon the rudimentary SummaRunner~\cite{summarunner,content_sel_summ} architecture with Convolutional Neural Network (CNN) based sentence encoders. We input the concatenated padded sequence of GloVe~\cite{glove} embeddings for tokens in a sentence to a two-layered CNN sentence vectorizer. Internally, a CNN layer applies a sequence of convolution and batch normalization followed by a Leaky ReLU activation. The output from the two stacked CNNs is then max pooled. Inside, three CNN stacks with kernel sizes of 3, 4, and 5 are used whose outputs are concatenated to form the sentence embedding.

Individual CNN-based sentence vectors are passed through a layer of bi-GRUs (bidirectional Gated Recurrent Units)~\cite{gru} that produce a sequence of contextual sentence embeddings. All such contextual embeddings are max pooled and then passed through a fully connected (FC) layer to derive a document embedding vector that captures the global context of the complete document. Each sentence score is then computed based on aspects such as their content, novelty, etc.,~\cite{summarunner} which are in turn expressed as functions of their contextual embedding and the document embedding vector. \\


\noindent\textbf{BERT$_{\scriptsize{\textnormal{Reg}}}$}:  
The use of Transformer-based~\cite{transformer} models such as BERT~\cite{bert} have yielded state-of-the-art results on various NLP tasks. We adapt \textsc{BertSumExt}'s~\cite{presumm} architecture of sentence classification to our sequential regression setting. To generate contextual sentence embeddings, we add \texttt{[SEP]} tokens to mark sentence ends and distinctive \texttt{[REG]} tokens at sentence beginnings. Downstream, the contextual embeddings of the \texttt{[REG]} tokens are used for sequence regression.

For each input token, its token, position, and segment embeddings are generated~\cite{bert}. Here, the segment embeddings allow differentiation between odd and even positioned sentences within BERT (since traditional BERT has been trained with only two segments). The token, position, and segment embeddings of the defined input tokens are passed to BERT, which produces a sequence of \textit{refined} token embeddings. Now, the sequence of BERT-based \textit{sentence} embeddings (embeddings corresponding to the \texttt{[REG]} tokens) is summed with their sinusoidal sentence positional embeddings and passed through a two-layer Transformer model to yield contextual sentence embeddings aware of multi-sentence discourse. \cite{bertsumm} showed the usefulness of adding a two-layer Transformer that captures inter-sentence relationships as opposed to directly using BERT-generated embeddings for sentence-sequence labeling. Finally, the generated contextual embeddings of the individual sentences are passed through a linear layer with a sigmoid activation to generate regression scores $\in$ $[0, 1]$. Fig.~\ref{fig:bert_reg} shows an overview of BERT$_{\scriptsize{\textnormal{Reg}}}$.

\section{Evaluation Metrics}
\label{sec:metrics}

We utilize the following automatic evaluation metrics to measure the performance of our models.

\begin{itemize}
\itemsep 0.6em
\item \textbf{Top $k$ overlap}\ \ Let $A_k$ and $P_k$ be the sets of actual and predicted top-$k$ highest scored sentences, respectively. We define top $k$ overlap (top$_k$) as $|A_k \cap P_k|/k$ and report its percentage. Thus, top$_1$ marks whether a method correctly identifies the highest scored sentence.

\item \textbf{Regression errors}\ \ We report Mean Squared Error (MSE) and Mean Absolute Error (MAE) between arrays of actual and predicted sentence labels.

\item \textbf{Rank Correlation Metrics}\ \ Top $k$ overlap focuses only on the identification of the highest scored sentences. To get a picture of how well various algorithms rank and order the complete set of sentences in a document, we use Spearman's rank correlation ($\rho$) and Kendall's Tau ($\tau$). $\rho, \tau \in [-1,1]$.

\item \textbf{nDCG}\ \ Top $k$ overlap, $\rho$, and $\tau$ only consider sentence ranking, while regression errors only account for sentence scores. Normalized Discounted Cumulative Gain (nDCG)~\cite{ndcg} captures the normalized gain or usefulness of a sentence based on both its position in the inferred rank list and its actual score. nDCG $\in$ $[0, 1]$.

Intuitively, the penalty for downscoring and inverting ranks of two truly high scored sentences should be lower than that of inverting ranks of two negligibly scored ones. 
Rank Correlation metrics do not take scores into account and fail to capture such insights. Whereas regression errors only focus on true and forecasted scores ignoring the score-induced rank ordering. Thus, broadly, nDCG is the most holistic metric for our task.


\end{itemize}

\setlength{\tabcolsep}{3pt}
\begin{table}[t]
\small
\centering
\caption{Sentence Popularity Forecasting Results}
\begin{tabular}{|P{1.2cm}|P{0.7cm}|P{0.8cm}P{0.8cm}P{0.8cm}|P{1cm}P{1cm}|P{1cm}P{1cm}|P{1cm}|}
\hline
Method&TL&Top$_1$&Top$_2$&Top$_3$&MSE&MAE&$\tau$&$\rho$&nDCG\\
\hline
\hline
Position&$-$&6.92&10.81&16.46&0.0079&0.0530&0.0334&0.0424&0.5804\\
TextRank&$-$&6.08&12.58&19.08&0.0316&0.1486&0.0345&0.0474&0.6313\\
LexRank&$-$&\textbf{18.76}&\textbf{29.95}&\textbf{37.84}&\textbf{0.0072}&\textbf{0.0503}&\textbf{0.0545}&\textbf{0.0705}&\textbf{0.7324}\\
\hline
\multirow{4}{*}{Base$_{\scriptsize{\textnormal{Reg}}}$}&$\times$&9.12&15.49&20.93&0.0083&\textbf{0.0452}&0.0534&0.0746&0.6228\\
&S$_1$&9.35&14.91&20.49&0.0075&0.0478&0.0428&0.0592&0.6323\\
&S$_2$&\textbf{10.08}&\textbf{16.63}&\textbf{22.83}&\textbf{0.0073}&0.0468&\textbf{0.0545}&\textbf{0.0751}&\textbf{0.6465}\\
&S$_{\scriptsize{\textnormal{L}}}$&9.16&14.85&20.76&0.0075&0.0475&0.0465&0.0638&0.6307\\
\hline
\multirow{4}{*}{BERT$_{\scriptsize{\textnormal{Reg}}}$}&$\times$&27.53&38.89&45.89&0.0055&0.0335&0.0704&0.0955&0.7921\\
&S$_1$&27.54&38.73&45.86&\textbf{\underline{0.0052}}&0.0342&\textbf{\underline{0.0734}}&\textbf{\underline{0.0988}}&0.8009\\
&S$_2$&\textbf{\underline{28.34}}&39.08&46.17&0.0053&0.0332&0.0646&0.0876&0.8007\\
&S$_{\scriptsize{\textnormal{L}}}$&28.11&\textbf{\underline{39.53}}&\textbf{\underline{46.96}}&0.0053&\textbf{\underline{0.0331}}&0.0510&0.0674&\textbf{\underline{0.8025}}\\
\hline
\end{tabular}
\label{tab:popforecast}
\end{table}

\section{Experimental Details}
\label{sec:experiments}
We used publicly available implementations of TextRank\footnote{\url{https://github.com/summanlp/textrank}} and LexRank\footnote{\url{https://github.com/crabcamp/lexrank}}. We normalized inferred LexRank scores to a $[0, 1]$ range (by dividing by the sum of all sentence scores) for computing regression errors. To create the auxiliary transfer learning subtasks, we utilized the non-anonymized version of CNN-DailyMail dataset~\cite{cnn_dm_hermann}. We used spaCy’s\footnote{\url{https://spacy.io/}} dependency parser and NLTK's\footnote{\url{https://www.nltk.org/}} \textit{sent\_tokenize} function during preprocessing.

We used a maximum sequence limit of $100$ sentences in the layer of bi-GRUs within our Base$_{\scriptsize{\textnormal{Reg}}}$ model. The CNN-sentence encoder had an upper input limit of $100$ tokens per sentence. For training our Base$_{\scriptsize{\textnormal{Reg}}}$ model, we used Adam optimizer with a batch size of $256$ documents, an initial learning rate of $10^{-5}$, with default hyperparameters. We used the $6$-layer bert-base-uncased model in our BERT$_{\scriptsize{\textnormal{Reg}}}$ experiments and considered a maximum sequence length of $1536$ tokens. Adam optimizer was used with a learning rate of $2.10^{-3}$, with other hyperparameters set to their default values. We split documents exceeding the maximum sequence length into overlapping sliding windows with a predefined maximum stride of $10$ sentences. During inference, we adopted the same splitting technique. For a sentence $s$ within the stride of two consecutive windows, we scored $s$ as the mean of the computed scores from the two windows.

We trained all of our models on NVIDIA RTX $2080$Ti(s). Base$_{\scriptsize{\textnormal{Reg}}}$ models were trained on a single GPU for a maximum of $4$ epochs with early stopping. A single epoch on our transfer learning task took over $6$ hours, while one epoch for popularity forecasting training took close to $45$ minutes. We trained BERT$_{\scriptsize{\textnormal{Reg}}}$ based models on $4$ GPUs for $50,000$ optimizer steps with early stopping turned off for both the popularity forecasting and the transfer learning tasks. The training time ranged between $10$ to $12$ hours.

\section{Results}
\label{sec:results}

In this section, we analyze the performance of baselines and our proposed approach for proactive sentence-specific popularity forecasting. We then share some insights on the interrelations between our primary task of sentence popularity forecasting and our auxiliary Transfer Learning (TL) task of salience prediction.

\begin{table}
\small
\centering
\caption{Performance of various approaches on auxiliary transfer learning subtasks (S$_1$, S$_2$, S$_{\scriptsize{\textnormal{L}}}$)}
\label{tab6}
\begin{tabular}{|P{0.6cm}|P{1.4cm}|P{0.8cm}P{0.8cm}P{0.8cm}|P{1cm}P{1cm}|P{1cm}P{1cm}|P{1cm}|}
\hline
Task&Approach&Top$_1$&Top$_2$&Top$_3$&MSE&MAE&$\tau$&$\rho$&nDCG\\
\hline
\hline
\multirow{5}{*}{S$_1$}&Position&12.86&26.30&34.04&0.0010&0.0215&\textbf{0.2030}&\textbf{0.2855}&0.8682\\
&TextRank&15.55&24.28&30.89&0.0243&0.1487&0.0771&0.1082&0.8999\\
&LexRank&13.14&21.91&28.45&0.0007&0.0178&0.0561&0.0798&0.8740\\
& Base$_{\scriptsize{\textnormal{Reg}}}$& 17.01& 26.60& 34.00& 0.0006& 0.0167& 0.1387& 0.1967& 0.8933\\
& BERT$_{\scriptsize{\textnormal{Reg}}}$& \textbf{26.41}& \textbf{36.64}& \textbf{43.20}& \textbf{0.0004}& \textbf{0.0130}& 0.1372& 0.1891& \textbf{0.9274}\\ 
\hline

\multirow{5}{*}{S$_2$}&Position&11.24&25.03&32.77&0.0044&0.0418&\textbf{0.1496}&\textbf{0.2101}&0.7113\\
&TextRank&9.29&17.59&23.43&0.0280&0.1541&0.0407&0.0578&0.6665\\
&LexRank&11.74&20.68&26.88&0.0045&0.0422&0.0473&0.0669&0.6847\\
& Base$_{\scriptsize{\textnormal{Reg}}}$& 17.19& 27.56& 35.83& 0.0041& 0.0373& 0.1391& 0.1989& 0.7382\\
& BERT$_{\scriptsize{\textnormal{Reg}}}$& \textbf{23.32}& \textbf{36.26}& \textbf{43.68}& \textbf{0.0034}& \textbf{0.0332}& 0.1108& 0.1559& \textbf{0.7946}\\ 
\hline

\multirow{5}{*}{S$_{\scriptsize{\textnormal{L}}}$}&Position&13.60&27.57&35.18&0.0009&0.0211&\textbf{0.2050}&\textbf{0.2881}&0.8760\\
&TextRank&11.72&20.05&26.77&0.0245&0.1488&0.0719&0.1016&0.8776\\
&LexRank&12.40&21.71&27.95&0.0007&0.0182&0.0546&0.0778&0.8657\\
& Base$_{\scriptsize{\textnormal{Reg}}}$& 15.13& 24.75& 32.36& 0.0007& 0.0175& 0.1385& 0.1966& 0.8780\\
& BERT$_{\scriptsize{\textnormal{Reg}}}$& \textbf{24.24}& \textbf{34.96}& \textbf{41.76}& \textbf{0.0005}& \textbf{0.0141}& 0.1329& 0.1847& \textbf{0.9152}\\
\hline

\end{tabular}
\end{table}

\subsection{Sentence-Specific Popularity Forecasting}
\subsubsection{Unsupervised sentence ranking}

Table~\ref{tab:popforecast} shows performance of unsupervised sentence ranking baselines on sentence-specific popularity forecasting. LexRank turns out the best unsupervised method beating Position and TextRank on all metrics.

\subsubsection{Proposed Methods}

We present results of our supervised models in Table~\ref{tab:popforecast}. As expected, BERT$_{\scriptsize{\textnormal{Reg}}}$ is the best architecture for the task. We see performance upgrades across the board upon employment of the Transfer Learning setup.
Though there is no consistent winner among the three Transfer Learning subtasks\footnote{We prefer TL = S$_{\scriptsize{\textnormal{L}}}$ as it outperforms others on Top$_k$ ($k$ = 2, 3), MAE, and most importantly, nDCG. A downstream application focused purely on accurate ranking could use TL = S$_{\scriptsize{\textnormal{1}}}$, while another requiring identification of the most popular sentence could use TL = S$_{\scriptsize{\textnormal{2}}}$.}, the best result for each metric is achieved using some form of transfer learning.
BERT$_{\scriptsize{\textnormal{Reg}}}$ with TL = S$_{\scriptsize{\textnormal{L}}}$ boosted average nDCG over 1\% from plain BERT$_{\scriptsize{\textnormal{Reg}}}$ and over 7\% from the best unsupervised baseline. We performed $t$-tests which showed that BERT$_{\scriptsize{\textnormal{Reg}}}$ with TL $=$ S$_{\scriptsize{\textnormal{L}}}$ significantly outperformed vanilla BERT$_{\scriptsize{\textnormal{Reg}}}$ on nDCG at significance level $p<0.01$. Though Base$_{\scriptsize{\textnormal{Reg}}}$ is ineffective for popularity forecasting, we observe the same trend of transfer learning as a positive addition. These results demonstrate that transfer learning from salience prediction significantly improves sentence popularity forecasting.

We attribute the performance enhancement to two factors. Firstly, datasets used for both tasks are sourced from online news documents, and transfer learning allows the model to witness more domain-specific data~\cite{domain_adapt}. Secondly, though popularity forecasting differs from salience prediction (Sec.~\ref{sec:pop_sal}), they have certain similarities, such as penalizing not lexically dense sentences or understanding that specific sentences do not carry any notable information. Our proposed approach grounds the network to capture such relationships.

\begin{table*}
\small
\centering
\caption{Cross-task evaluation $-$ performance of BERT$_{\scriptsize{\textnormal{Reg}}}$ trained for popularity forecasting (PF) evaluated on salience prediction and vice-versa}
\label{tab:5}
\begin{tabular}{|P{0.7cm}|P{0.6cm}|P{0.8cm}P{0.8cm}P{0.8cm}|P{1cm}P{1cm}|P{1cm}P{1cm}|P{1cm}|}
\hline
Train& Eval&Top $1$&Top $2$&Top $3$&MSE&MAE&$\tau$&$\rho$&nDCG\\
\hline 
\hline
S$_1$&\multirow{3}{*}{PF}&10.97&18.31&25.69&0.0068&0.0475&0.0430&0.0598&0.6864\\
S$_2$&&10.74&19.83&27.89&0.0077&0.0455&0.0380&0.0531&0.6942\\
S$_{\scriptsize{\textnormal{L}}}$&&11.44&19.09&27.02&0.0068&0.0476&0.0428&0.0595&0.6937\\
\hline
\multirow{3}{*}{PF}&S$_1$&8.36&16.63&22.64&0.0020&0.0301&0.0600&0.0847&0.8603\\
&S$_2$&8.51&16.34&22.53&0.0053&0.0430&0.0373&0.0525&0.6579\\
&S$_{\scriptsize{\textnormal{L}}}$&9.07&16.62&22.86&0.0020&0.0304&0.0563&0.0798&0.8524\\
\hline
\end{tabular}
\end{table*}

\subsection{Popularity and Salience}
\label{sec:pop_sal}
Salient sentences effectively capture summary-inclusion worthy ideas that are central to the core semantics of an article~\cite{text_match}. Unlike salient information bits, an information piece might deviate significantly from an article's primary topic yet be popular. Consider the following sentence from a particular article\footnote{\url{https://www.npr.org/2018/07/03/625581627/another-top-justice-department-lawyer-steps-down-following-earlier-departures}} (with ID 34499) within \textsc{InfoPop}: `Weinsheimer has spent 27 years at DOJ, where he tried homicide and public corruption cases.' The sentence is not salient enough to be included in a summary as it is barely related to the article's core topic (Scott Schools' resignation). But, it contains one of the most popular information pieces in the document.

\subsubsection{Quantitative analysis}
We evaluate our unsupervised baselines and supervised neural models on the transfer learning subtasks (Table~\ref{tab6}). The position baseline achieves the best rank correlation scores, explainable based on the pyramid structure of news reports. BERT$_{\scriptsize{\textnormal{Reg}}}$ beats other approaches on nDCG and in identifying the most salient sentences (Top $k$).

Comparing with Table~\ref{tab:popforecast}, we see how sentence ranking baselines are more capable of capturing information salience than forecasting information popularity. Position baseline's scores for salience prediction are significantly greater than for popularity forecasting, indicating that the salience of the initial sentences in news articles is typically higher than their popularity.

Comparison with Table~\ref{tab:popforecast} shows that supervised salience prediction models achieve much better results on $\rho$, $\tau$, and nDCG compared to popularity forecasting models utilizing the same underlying architecture. Also, note how Base$_{\scriptsize{\textnormal{Reg}}}$, which performed poorly on popularity forecasting, achieves respectable scores for salience prediction. These experiments identify salience prediction as the less complicated problem, i.e., given a document, it is easier to compute how central and summary-worthy a contained information piece is than forecasting its popularity. Table~\ref{tab:data_InfoPop} shows an excerpt from an \textsc{InfoPop} article with actual and forecasted popularity plus predicted salience labels for sentences showcasing the variation between information popularity and salience within the same document.

\subsubsection{Empirical cross-task evaluation}
\label{sec:cross_task}
We performed an empirical cross-task analysis~\cite{cross_task_spanrep,cross_task_query_based} to capture the degree of relatedness between primary and auxiliary tasks.
We used the BERT$_{\scriptsize{\textnormal{Reg}}}$ popularity forecasting model to infer sentences' labels for documents in the salience-prediction dataset. We evaluated these scores against actual sentence salience labels of the three types -- S$_1$, S$_2$, and S$_{\scriptsize{\textnormal{L}}}$. Similarly, we evaluated models trained on S$_1$, S$_2$, and S$_{\scriptsize{\textnormal{L}}}$ upon our primary popularity forecasting task. We show results in Table~\ref{tab:5}. Predictably, we observe a massive drop in performance when we switch to the cross-task setting. Values across all metrics fall below those achieved by some unsupervised ranking baseline. This further experimentally showcases the distinction between information popularity and information salience.

\section{Conclusion}
\label{sec:conclusion}
In this work, we introduced the task of proactively forecasting sentence-specific information popularity. We contribute \textsc{InfoPop}, a dataset containing 51,770 news articles from 26 news websites with over 1.7 million sentences labeled with normalized popularity scores. We experimented with both unsupervised and supervised baselines and demonstrated the efficacy of our novel STILTs-based Transfer Learning approach involving an auxiliary supervised task of salience prediction. Our best models achieved nDCG values over 0.8 for sentence-specific popularity forecasting. Our analysis presents an interesting takeaway: though popularity forecasting and salience prediction are quite different problems, transferring the learning from a salience prediction task enhances a model's popularity forecasting proficiency.
In future, we aim to explore potential business applications of sentence popularity forecasting in problems such as pull quote extraction, popularity-guided text summarization, etc. We also plan on exploring a multi-task learning approach considering the popularity forecasting and salience prediction problems.

\bibliographystyle{ACM-Reference-Format}
\bibliography{bibfile}

\end{document}